\documentclass[11pt,a4paper]{article}
\usepackage[hyperref]{acl2021}
\usepackage{times}
\usepackage{latexsym}

\usepackage{graphicx}
\usepackage{microtype}
\usepackage{url}
\usepackage{multirow}
\usepackage{subcaption}
\usepackage{arydshln}
\usepackage{amssymb}
\usepackage{adjustbox}
\usepackage{soul}
\usepackage{color}

\colorlet{soulred}{red!50}

\colorlet{soulbleu}{cyan!20}

\colorlet{soulgreen}{green!20}
\DeclareRobustCommand{\hlgreen}[1]{{\sethlcolor{soulgreen}\hl{#1}}}

\colorlet{soulyellow}{yellow!40}

\colorlet{soulorange}{orange!30}
\DeclareRobustCommand{\hlorange}[1]{{\sethlcolor{soulorange}\hl{#1}}}

\colorlet{soulpurple}{blue!50}

\aclfinalcopy 

\title{Improve Query Focused Abstractive Summarization by \\ Incorporating Answer Relevance}

\author{Dan Su\thanks{$^*$ The two authors contribute equally.} \,, Tiezheng Yu$^*$, Pascale Fung \\
Center for Artificial Intelligence Research (CAiRE)\\
Department of Electronic and Computer Engineering\\
The Hong Kong University of Science and Technology, Clear Water Bay, Hong Kong\\
\texttt{\{dsu,tyuah\}@connect.ust.hk},  \texttt{pascale@ece.ust.hk}}

\date{}
\begin{document}
\maketitle
\begin{abstract}
Query focused summarization (QFS) models aim to generate summaries from source documents that can answer the given query. Most previous work on QFS only considers the query relevance criterion when producing the summary. However, studying the effect of answer relevance in the summary generating process is also important. In this paper, we propose QFS-BART, a model that incorporates the explicit answer relevance of the source documents given the query via a question answering model, to generate coherent and answer-related summaries. Furthermore, our model can take advantage of large pre-trained models which improve the summarization performance significantly. Empirical results on the Debatepedia dataset show that the proposed model achieves the new state-of-the-art performance.\footnote{The code is released at: \url{https://github.com/HLTCHKUST/QFS}}
\end{abstract}

\section{Introduction}

\begin{table}[t]
    \centering
    \begin{adjustbox}{width={0.45\textwidth},totalheight={\textheight},keepaspectratio}
\begin{tabular}{|p{1\columnwidth}|}
\hline
\textbf{Document}: Interrogator Ali Soufan said in an April op-ed article in the New York Times: ``It is inaccurate to say that Abu Zubaydah had been uncooperative [and that enhanced interrogation techniques supplies interrogators with previously unobtainable information]. Along with another f.b.i. agent and with several c.i.a. officers present I questioned him from March to June before the harsh techniques were introduced later in August. \hlorange{Under traditional interrogation methods he provided us with important actionable intelligence.}'' \\ \hline
\textbf{Query}: Are traditional interrogation methods insufficient?                                                                                             \\ \hline
\textbf{Summary}: The same info can be obtained by traditional interrogations.                                                                                    \\ \hline
\end{tabular}
    \end{adjustbox}
    \caption{An example of QFS. The input is a document and a corresponding query, and the \hlorange{highlight} sentence is the answer from our QA module. We observe that the summary and the answers are very correlated.}
    \label{tab:introduction}
\end{table}

Query focused summarization (QFS) models aim to extract essential information from a source document(s) and organize it into a summary that can answer a query~\cite{dang2005overview}. The input can be either a \textit{single} document that has multiple views or \textit{multiple} documents that contain multiple topics, and the output summary should be focused on the given query. QFS has various applications (e.g., a personalized search engine that provides the user with an overview summary based on their query~\cite{su2020caire}).


Early work on the QFS task mainly focused on generating extractive summaries~\cite{davis2012occams, daume2006bayesian, feigenblat2017unsupervised, xu2020coarse}, which may contain unreadable sentence ordering and lack cohesiveness. Other work on abstractive QFS incorporated the query relevance into existing neural summarization models~\cite{nema2017diversity,baumel2018query}. The closest work to ours was done by~\cite{su2020caire} and~\cite{xu2020abstractive, xu2020coarse}, who leveraged an external question answering (QA) module in a pipeline framework to take into consideration the answer relevance of the generated summary. However, they only used QA as distant supervision to retrieve relevant segments for generating the summary, but did not take into consideration the answer relevance in the generation model. As shown in the Table~\ref{tab:introduction}, the query focused summary is correlated to the answer extracted from the QA module.

On the other hand, recent neural summarization models~\cite{paulus2017deep, gehrmann2018bottom, zhang2020pegasus} have achieved remarkable performance in~\textit{generic} abstractive summarization by taking advantage of large pre-trained language models~\cite{lewis2019bart,zhang2020pegasus}. Yet, how to leverage these models and adapt them to the QFS task remains unexplored.

In this work, we propose QFS-BART, a BART-based~\cite{lewis2019bart} framework for abstractive QFS that incorporates explicit answer relevance. We leverage a state-of-the-art QA model~\cite{su2019generalizing} to predict the answer relevance of the given source documents to the query, then further incorporate the answer relevance into the BART-based generation model. We conduct empirical experiments on the Debatepedia dataset, one of the first large-scale QFS datasets~\cite{nema2017diversity}, and achieve the new state-of-the-art performance on the ROUGE metrics compared to all previously published work.


Our contributions in this work are threefold:
\begin{itemize}
\item Our work demonstrates the effectiveness of the answer relevance score in neural abstractive QFS.
\item We propose an effective method to incorporate the answer relevance score into the pre-trained language models which can produce more query-relevant summaries.
\item Our model reaches the state-of-the-art performance on a single-document QFS dataset (Debatepedia), and brings substantial improvements over several strong baselines on two multi-document QFS datesets (DUC 2006, 2007).
\end{itemize}

\section{Related Work}
Abstractive summarization models aim to generate short, concise and readable text that extracts the salient information from a document. In the past few years, significant achievements~\cite{see2017get, liu2019text, lewis2019bart, dong2019unified} have been made in the \textit{generic} abstractive summarization task which is attributed to the advanced neural architectures and the availability of large-scale datasets~\cite{sandhaus2008new, hermann2015teaching, grusky2018newsroom}. 

QFS is a more complex task that aims to generate a summary according to the query and its relevant document(s).  \citet{nema2017diversity} proposed an encode-attend-decode system with an additional query attention mechanism and diversity-based attention mechanism to generate a more query-relevant summary. \citet{baumel2018query} incorporated query relevance into a pre-trained abstractive summarizer to make the model aware of the query, while~\citet{xu2020abstractive} discovered a new type of connection between generic summaries and QFS queries, and provided a universal representation for them which allows \textit{generic} summarization data to be further exploited for QFS. ~\citet{su2020caire}, meanwhile, built a query model for paragraph selection based on the answer relevance score and iteratively summarized paragraphs to a budget. Although ~\citet{xu2020abstractive} and ~\citet{su2020caire} utilized QA models for sentence- or paragraph- level answer evidence ranking, they did not make use of answer relevance to query-forcused generation.

To the best of our knowledge, we are the first to leverage explicit answer relevance to abstractive QFS. In addition, our approach can be easily combined with pre-trained Transformers ~\cite{song2019mass, dong2019unified, lewis2019bart, xiao2020ernie}, which have shown great success for the \textit{generic} abstractive summarization task. 

\section{Methodology}

\begin{figure}
    \centering
    \includegraphics[scale=0.7]{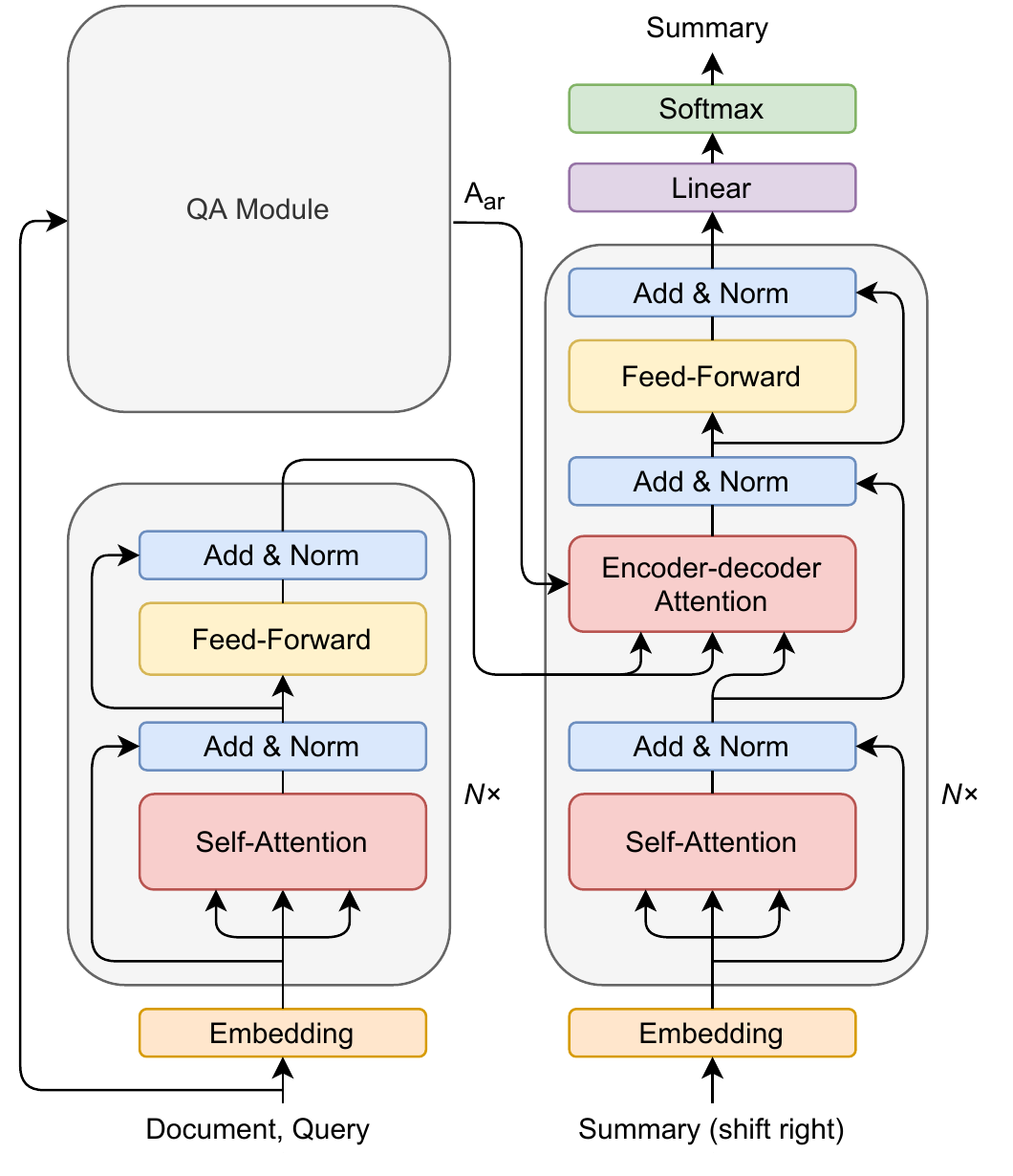}
    \caption{The framework of QFS-BART. The QA module calculates the answer relevance scores, and we incorporate the scores as explicit answer relevance attention to the encoder-decoder attention.}
    \label{fig:QFS-BART-model}
\end{figure}

In this section, we present our approach to incorporating the answer relevance into QFS. First, we introduce the method of generating answer relevance scores. Then, we describe our answer relevance attention in the Transformer-based model. Third, we introduce our QFS-BART model in which the decoder is composed of a stack of answer relevance decoding layers, as shown in Figure \ref{fig:QFS-BART-model}.

\subsection{Answer Relevance Generation}
In recent years, neural models~\cite{yang2019end, su2019generalizing} have shown remarkable achievements in QA tasks. In order to apply QA models to the QFS task, we use HLTC-MRQA~\cite{su2019generalizing} to generate the answer relevance score for each word in context. The reason for choosing HLTC-MRQA is twofold: 1) it shows robust generalization and transferring ability on different datasets, and 2) the model shows great performance in QA tasks and significantly outperforms the BERT-large baseline by a large margin. The HLTC-MRQA is introduced as follows.

Based on XLNet~\cite{yang2019xlnet}, HLTC-MRQA is fine-tuned on multiple QA datasets with an additional multilayer perceptron (MLP). Given a context that contains $n$ words, the model outputs a distribution $s \in (0, 1)$ for each word's probability of being the start word of the answer and a probability distribution $e \in (0, 1)$ to be the end word of answer. To generate the answer relevance score $r$ for each word, we calculate it by summing two distributions:
\begin{equation}
    r = s + e,
\end{equation}
where $r \in (0, 2)$.

\subsection{Answer Relevance Attention}
Scaled dot-product attention~\cite{vaswani2017attention} is the core-component of the Transformer-based model:
\begin{equation}
    Attention(Q, K, V) = softmax( \frac{QK^{T}}{ \sqrt{d}}V),
\end{equation}
where $d$ is the dimension of the query matrix $Q$, key matrix $K$ and value matrix $V$. The Transformer encoder is constructed by self-attention layers, where all of the keys, values and queries come from the input sequence. This makes each token in the input attend to all other tokens. The Transformer decoder layer is a combination of a self-attention layer and encoder-decoder attention layer. In the encoder-decoder attention layer, the query comes from the decoder's self-attention layer, and the key and value come from the output of the encoder. This allows every generated token to attend to all tokens in the input sequence.

In this work, we propose to incorporate the word-level answer relevance score as additional explicit encoder-decoder attention in the transformer decoder. Given a document with $n$ tokens, we generate a summary with a maximum length of $m$ tokens. Let $x^{l} \in \mathbb{R}^{n*d}$ denotes the output of the $l$-th transformer encoder layer and $y^{l} \in \mathbb{R}^{m*d}$ denotes the output of the $l$-th transformer decoder layer's self-attention layer. The encoder-decoder attention $\alpha^{l} \in \mathbb{R}^{m*n}$ can be computed as:
\begin{equation}
    \alpha^{l} = softmax(\frac{(y^{l}W_{Q})(x^{l}W_{K})}{\sqrt{d_{k}}} + A_{ar}),
\end{equation}
where $W_{Q}$ and $W_{K} \in \mathbb{R}^{d_{k}*d_{k}}$ are parameter weights and $A_{ar} \in \mathbb{R}^{m*n}$ is our explicit answer relevance score. Since the original answer relevance score is an $n$-dimensional vector, we repeat it $m$ times to generate an $m$ by $n$ attention matrix, which means our answer relevance attention is equal to all generated tokens.

\subsection{QFS-BART}

\begin{table*}[ht]
    \centering
    \begin{adjustbox}{width={0.7\textwidth},totalheight={\textheight},keepaspectratio}
    \begin{tabular}{lccc}
    \hline
    \textbf{Models}                                     & \textbf{ROUGE-1} & \textbf{ROUGE-2} & \textbf{ROUGE-L} \\ \hline \hline
    \multicolumn{4}{l}{\qquad \textit{Without Pre-training}}  \\
    Transformer~\cite{vaswani2017attention}             & 28.16 & 17.48 & 27.28 \\
    Transformer (CONCAT)                                & 41.72 & 33.62 & 41.25 \\
    Transformer (ADD)                                   & 41.10 & 33.35 & 40.72 \\
    SD2*~\cite{nema2017diversity}                      & 41.26 & 18.75 & 40.43 \\ 
    CSA Transformer*~\cite{xie2020conditional}          & 46.44 & 37.38 & 45.85 \\ \hdashline
    \multicolumn{4}{l}{\qquad \textit{With Pre-training}}  \\ 
    RSA Word Count*~\cite{baumel2018query}              & 53.09 & 16.10 & 46.18 \\
    QR-BERTSUM-TL*~\cite{laskar2020query}               & 57.96 & \textbf{45.20} & 57.05 \\ \hline
    \textbf{BART-FT}                                    & 57.98 & 43.62 & 56.30 \\
    \textbf{QFS-BART}                                   & \textbf{59.02} & 44.59 & \textbf{57.44} \\ \hline
    \end{tabular}
    \end{adjustbox}
    \caption{ROUGE-F1 scores for Debatepedia QFS dataset. Results with * mark are taken from the corresponding papers. The previous work can be divided into two categories: 1) training the models from scratch, and 2) using pre-trained models and fine-tuning on a QFS dataset.}
    \label{tab:main_results}
\end{table*}

Generative pre-trained models~\cite{dong2019unified, lewis2019bart, raffel2019exploring} have shown remarkable performance in natural language generation (NLG), including text summarization. We choose to combine our answer relevance attention with BART ~\cite{lewis2019bart}, a denoising autoencoder built with a sequence-to-sequence model, for two reasons: 1) BART achieves state-of-the-art performance on several summarization datasets (i.e. CNN/DailyMail ~\cite{hermann2015teaching} and Xsum ~\cite{narayan2018don}). 2) BART follows the standard Transformer encoder-decoder architecture, and we can easily combine the answer relevance as explicit attention to the encoder-decoder attention layers. In detail, we incorporate the same answer relevance attention for all Transformer decoder layers.

Domain adaption for natural language processing tasks is widely studied ~\cite{blitzer2007biographies, daume2009frustratingly, liu2020crossner, yu2021adaptsum}. ~\citet{hua2017pilot} first studied the adaptation of neural summarization models and showed that the models were able to select salient information, even when trained on out-of-domain data. Inspired by this, we leverage a two-stage fine-tuning method for our QFS-BART. In the first stage, we directly fine-tune the original BART model with the Xsum dataset, and in the second stage, we fine-tune our QFS-BART model with QFS datasets. All the parameters in the model are initialized from the first stage. In order to make the model capture both query relevance and answer relevance, the input text is formatted in the following way:

\centerline{[CLS] document [SEP] query.}
The answer relevance attention score for the document is generated by the QA model, and we take the maximum number in the document as the attention score for all the words in the query.

\section{Experimental Setup}
\paragraph{Datasets} We use multiple QA datasets, including SQuAD~\cite{rajpurkar2016squad}, NewsQA~\cite{trischler2016newsqa}, TriviaQA~\cite{joshi2017triviaqa}, SearchQA~\cite{dunn2017searchqa}, HotpotQA~\cite{yang2018hotpotqa} and NaturalQuestions~\cite{kwiatkowski2019natural} to train HLTC-MRQA, following~\citet{su2019generalizing}. We evaluate our model on the Debatepedia dataset~\cite{nema2017diversity} and DUC2005-7 dataset (in Appendix).

\paragraph{Training Details}
For all the experiments, we use the BART-large version to implement our models. We use a mini-batch size of 32 and train all the models on one V100 16G. During decoding, we use beam search with the beam size of 4. We decode until an end-of-sequence token is emitted and early stop when the generated summary reaches to 48 tokens.

\begin{table}[!ht]
    \centering
    \begin{adjustbox}{width={0.45\textwidth},totalheight={\textheight},keepaspectratio}
\begin{tabular}{p{1\columnwidth}}
\hline
\textbf{Document}: Interrogator Ali Soufan said in an April op-ed article in the New York Times: ``It is inaccurate to say that Abu Zubaydah had been uncooperative [and that enhanced interrogation techniques supplies interrogators with previously unobtainable information]. Along with another f.b.i. agent and with several c.i.a. officers present I questioned him from March to June before the harsh techniques were introduced later in August. Under traditional interrogation methods he provided us with important actionable intelligence.'' \\
\textbf{Query}: Are traditional interrogation methods insufficient?                                                                                             \\ \hline

\textbf{BART-FT}: Al Qaeda detainee Abu Zubaydah has been cooperative under traditional interrogation.                                                          \\ \hline

\textbf{QFS-BART}: \hlorange {The same info can be obtained by traditional interrogation.}                                                                                  \\ \hline

\textbf{Gold}: \hlgreen{The same info can be obtained by traditional interrogations.}                                                                                    \\ \hline
\end{tabular}
 \end{adjustbox}
    \caption{A example taken from Debatepedia test set. The generated summary from \textbf{QFS-BART} is almost the same as the gold summary.}
    \label{tab:case-study}
\end{table}

\section{Results \& Analysis}

We compare our proposed QFS-BART model with the following models: 1) \textbf{Transformer} does not consider the queries in the Debatepedia dataset. 2) \textbf{Transformer (CONCAT)} concatenates the query and the document. 3) \textbf{Transformer (ADD)} adds the query encoded vector to the document encoder. 4) \textbf{SD2} adds a query attention model and a new diversity-based attention model to the encode-attend-decode paradigm. 5) \textbf{CSA Transformer} combines conditional self-attention (CSA) with Transformer. 6) \textbf{RAS Word Count} incorporates query relevance into a pre-trained abstractive summarization model. 7) \textbf{QR-BERTSUM-TL} presents a transfer learning technique with the Transformer-based BERTSUM  model~\cite{liu2019text}. 8) \textbf{BART-FT} concatenates the document and query, and directly fine-tunes on the Debatepedia dataset.

We adopt ROUGE score~\cite{lin2004rouge} as the evaluation metric. As shown in Table \ref{tab:main_results}, QFS-BART significantly outperforms the models without pre-training. Compared with the models utilizing pre-training, ours improves the ROUGE-1 and ROUGE-L scores by a large margin.

\subsection{Case Study}
We present a case study comparing between the strong baseline BART-FT model, our QFS-BART model and the gold summary, shown in Table \ref{tab:case-study}. It's clear that the baseline model tends to copy spans from the document which are not directly related to the query and the QFS-BART model produces a more query- and answer- related summary.

\section{Conclusions}
In this work, we propose QFS-BART, an abstractive summarization model for query focused summarization. We use a generalizing QA model to make explicit answer relevance scores for all words in the document and combine them to the encoder-decoder attention. We also leverage pre-trained model (e.g. BART) and two-stage fine-tuning method which further improve the summarization performance significantly. Experimental results show the proposed model achieves state-of-the-art performance on Debatepedia dataset and outperforms several comparable baselines on DUC 2006-7 datasets.

\section*{Acknowledgments}
We want to thank the anonymous reviewers for their constructive feedback. This work is partially funded by ITS/353/19FP of the Innovation Technology Commission, the Hong Kong SAR Government.

\bibliographystyle{acl_natbib}
\bibliography{acl2021.bib}

\appendix

\section{Adapting QFS-BART to DUC 2005-7}
DUC 2005-7 are datasets for the multi-document query focused summarization (QFS) task. As shown in the Table \ref{tab:length_of_dataset}, the documents and summaries of the DUC datasets are extremely longer than those in the Debatepedia \cite{nema2017diversity} dataset. We thus need to adapt the QFS-BART model to handle the multi-document scenario and produce longer output.

\begin{table}[!ht]
    \centering
    \begin{adjustbox}{width={0.43\textwidth},totalheight={\textheight},keepaspectratio}
    \begin{tabular}{l|ccc}
    \hline
    \textbf{Datasets}   & \textbf{Document(s)}      & \textbf{Query}        & \textbf{Summary}  \\ \hline
    Debatepedia         & 66.40                     & 11.16                 & 9.97              \\
    DUC 2005            & 20058.12                  & 26.60                 & 243.56            \\   
    DUC 2006            & 14330.14                  & 23.30                 & 246.84            \\
    DUC 2007            & 10759.17                  & 21.57                 & 243.94            \\ \hline
    \end{tabular}
    \end{adjustbox}
    \caption{Average length of the input documents, queries and output summaries for the Debatepedia and DUC 2005-7 datasets. For the DUC datasets, we add up the lengths of all the documents related the same query}
    \label{tab:length_of_dataset}
\end{table}

In this paper, we introduce a two-step architecture: 1) Retrieve answer-related sentences given the query, rank them by the confidence score (generated from Equation \ref{equ: confidence_score}) and concatenate them. 2) Use our QFS-BART to produce an abstractive summary.

\subsection{Answer Retrieving}
We split documents into paragraphs and feed each paragraph to the QA model to get answer-related sentences. Then the sentences are ranked by the confidence score.
\paragraph{Document Segmentation}
The QA model selects one answer span given an input document, and the sentences that contain the span will be chosen as the answer-related sentences. Since we only retain the answer-related sentences as input to the next step, we set the maximum paragraph length to 300 words to avoid missing too much information in this step. Specifically, we feed text to the paragraph sentence by sentence until it reaches the maximum length. 

\paragraph{Answer Relevance Ranking}
The paragraphs are fed to the QA model to generate answer-related sentences and the corresponding answer relevance scores. We align each sentence with a confidence score from the corresponding answer span. The confidence score is defined as:

\begin{equation}
    confidence\_score = P_{start}+P_{end},
    \label{equ: confidence_score}
\end{equation}

where $P_{start}$ and $P_{end}$ is two probability distributions over the tokens in the context. $P_{start}(i)$/$P_{end}(i)$ the probability of the $i$-th token is the start/end of the answer span in context.

\subsection{Summary Generation}
We use the answer-related sentences and their answer relevance scores as the input to the QFS-BART model. The DUC 2005 dataset is used as a development set to optimize the model, and we evaluate the performance on the DUC 2006-7 dataset. We compare our QFS-BART with the following models.

\begin{table}[!ht]
    \centering
    \begin{adjustbox}{width={0.43\textwidth},totalheight={\textheight},keepaspectratio}
    \begin{tabular}{l|cccccc}
    \hline
    \multicolumn{1}{l|}{\multirow{2}{*}{\textbf{Models}}} & \multicolumn{3}{c}{\textbf{DUC 2006}}                    & \multicolumn{3}{c}{\textbf{DUC 2007}}                    \\ \cline{2-7} 
    \multicolumn{1}{c|}{}                                 & \textbf{1} & \textbf{2} & \textbf{SU4} & \textbf{1} & \textbf{2} & \textbf{SU4} \\ \hline
    \textbf{LEAD}                                         & 32.1             & 5.3              & 10.4               & 33.4             & 6.5              & 11.3               \\ 
    \textbf{TEXTRANK}                                     & 34.2             & 6.4              & 11.4               & 35.8             & 7.7              & 12.7               \\
    \textbf{HLTC-MRQA}                                    & 39.1             & 8.3              & 13.5               & \textbf{40.6}             & 9.6              & 14.7               \\ 
    \textbf{BART-CAQ}*                                    & 38.3             & 7.1              & 12.9               & 40.5             & 9.2              & 14.4               \\ \hline
    \textbf{BART-FT}                                      & 38.9             & 8.5              & 13.9               & 40.4             & \textbf{10.0}             & \textbf{15.1}               \\
    \textbf{QFS-BART}                                     & \textbf{39.4}             & \textbf{8.6}              & \textbf{14.1}               & 39.22              & 9.39              & 14.34                \\ \hline
    \end{tabular}
    \end{adjustbox}
    \caption{ROUGE-F1 scores for DUC 2006-7 dataset. Results with * mark are taken from the corresponding papers.}
    \label{tab: duc_results}
\end{table}

\paragraph{LEAD.} \cite{xu2020query} returns all leading sentences of the most recent document up to 250 words.

\paragraph{TEXTRANK.} \cite{mihalcea2004textrank} is a graph-based ranking model that incorporate two unsupervised methods for keyword and sentence extraction.

\paragraph{HLTC-MRQA.} truncates the ranked answer related sentences from our first step as the extractive summary.

\paragraph{BART-CQA.} \cite{su2020caire} uses QA models for paragraph selection and iteratively summarizes paragraphs to 250 words.

We adopt ROUGE-F1 score \cite{lin2004rouge} as the evaluation metric. As shown in Table \ref{tab: duc_results}, HLTC-MRQA significantly outperforms the LEAD and TEXTRANK baselines, which indicates the effectiveness of our answer retrieval. However, QFS-BART does not perform well on DUC 2006-7 datasets. We conjecture that the model can not converge to the task well with limited training samples (DUC 2005 contains only 300 samples).

\end{document}